\documentclass[letterpaper, 10 pt, conference]{ieeeconf}
\IEEEoverridecommandlockouts
\usepackage[utf8]{inputenc}

\usepackage{caption}

\usepackage[style=ieee,url=false,doi=false,isbn=false]{biblatex}
\AtEveryBibitem{\clearfield{month}}
\AtEveryBibitem{\clearfield{day}}
\AtEveryBibitem{
    \clearfield{urlyear}
    \clearfield{urlmonth}
}
\usepackage{xcolor}
\usepackage{booktabs}
\usepackage{layouts}
\usepackage[notext]{stix}
\usepackage{amsmath}
\renewcommand{\vec}{\boldsymbol}
\usepackage{mathtools}
\DeclarePairedDelimiter{\norm}{\lVert}{\rVert}

\usepackage{siunitx}
\usepackage[hidelinks]{hyperref}
\usepackage[capitalise]{cleveref}
\crefname{figure}{Fig.}{Figs.}
\Crefname{figure}{Fig.}{Figs.}
\crefname{table}{Tab.}{Tabs.}
\Crefname{table}{Tab.}{Tabs.}
\crefname{section}{Sec.}{Secs.}
\crefname{Section}{Sec.}{Secs.}
\crefname{equation}{}{}
\Crefname{equation}{Equation}{Equations}

\usepackage{tabularx}

\title{\LARGE \bf
Vision Foundation Models for Domain Generalisable\\ Cross-View Localisation in Planetary Ground--Aerial Robotic Teams
}
\author{Lachlan Holden$^{1}$, Feras Dayoub$^{1}$, Alberto Candela$^{2}$, David Harvey$^{3}$, and Tat-Jun Chin$^{1}$
}


\begin{document}

\makeatletter
\twocolumn[{%
\begin{@twocolumnfalse}
\maketitle
\begin{center}
\includegraphics[width=0.85\textwidth]{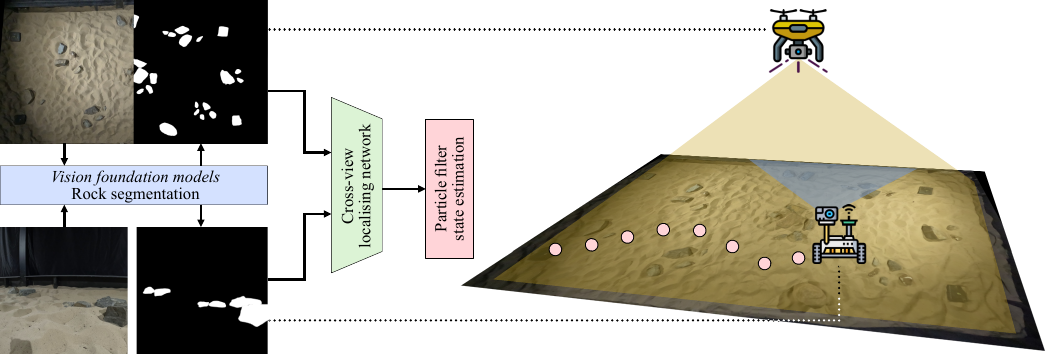}
\end{center}
\captionof{figure}{A high-level overview outlining our method of estimating the state of a rover within an aerial image.\vspace{1em}}
\label{fig:hero}
\end{@twocolumnfalse}
}]
{
  \footnotetext{$^{1}$AI for Space Group and $^{3}$Andy Thomas Centre for Space Resources, The University of Adelaide.
{\tt\small \{lachlan.holden,\allowbreak{}feras.dayoub,\allowbreak{}david.harvey,\allowbreak{}tat-jun.chin\}\allowbreak{}@adelaide.edu.au}}
  \footnotetext{$^{2}$Jet Propulsion Laboratory, California Institute of Technology, Pasadena, CA 91109, USA. {\tt\small alberto.candela.garza@\allowbreak{}jpl.nasa.gov}}
  \footnotetext{$^{\dagger}$Portions of this research were conducted by the Jet Propulsion Laboratory, California Institute of Technology, under a contract with the National Aeronautics and Space Administration (80NM0018D0004).}
}

\makeatother

\thispagestyle{empty}
\pagestyle{empty}

\begin{abstract}
Accurate localisation in planetary robotics enables the advanced autonomy required to support the increased scale and scope of future missions. The successes of the Ingenuity helicopter and multiple planetary orbiters lay the groundwork for future missions that use ground--aerial robotic teams. In this paper, we consider rovers using machine learning to localise themselves in a local aerial map using limited field-of-view monocular ground-view RGB images as input. A key consideration for machine learning methods is that real space data with ground-truth position labels suitable for training is scarce. In this work, we propose a novel method of localising rovers in an aerial map using cross-view-localising dual-encoder deep neural networks. We leverage semantic segmentation with vision foundation models and high volume synthetic data to bridge the domain gap to real images. We also contribute a new cross-view dataset of real-world rover trajectories with corresponding ground-truth localisation data captured in a planetary analogue facility, plus a high volume dataset of analogous synthetic image pairs. Using particle filters for state estimation with the cross-view networks allows accurate position estimation over simple and complex trajectories based on sequences of ground-view images.
\end{abstract}


\section{Introduction}
Accurate localisation remains a critical challenge for autonomous planetary rovers, especially without GPS in space. With the success of helicopter-and-orbiter-based imaging, there is growing potential to localise rovers using ground-view images and pre-collected aerial maps. Accurate localisation is a key part of growing rover autonomy, which is required to support future planetary exploration missions. In this work, we consider absolute map-based cross-view localisation for rovers using a monocular RGB camera with a constrained \ang{90} horizontal field of view.

RGB cameras are near-ubiquitous on planetary robotics platforms, and so vision-based localisation is of particular interest. Visual odometry is a common approach, but this, like other forms of odometry used for localisation, is subject to drift. Map-based localisation can instead provide an absolute form of localisation based on landmarks or features, and so is not subject to the same drift over time.

Cross-view localisation is a problem that has gained attention in both terrestrial and planetary contexts. On Earth, deep learning methods have shown success in localising car-collected ground-view images within satellite imagery \autocite{zhang.etal.2024_mtgeo,shugaev.etal.2024_arcgeo,zhu.etal.2022_transgeo}. Similar methods have been applied to the space domain \autocite{chen.etal.2024_metric,zhao.etal.2024_lunar}, but due to the limited availability of ground-truth labelled space data suitable for training of machine learning methods, they rely primarily on synthetic images.

A problem with relying on synthetic images for training is that they have have different appearance characteristics than real images. Machine learning models are typically sensitive to this and fail to generalise to real data if trained only on synthetic data. This problem is known as the ``domain gap''.

Our contributions in this work, as outlined in \cref{fig:hero}, are as follows: we introduce a new cross-view dataset of real and synthetic images in a planetary-analogue setting; we leverage semantic segmentation of rocks using vision foundation models to bridge the synthetic-real domain gap; we adapt a dual-encoder cross-view localisation network and train it with synthetic data; and we demonstrate accurate localisation with these methods using particle filters and real-world rover data.

\begin{figure*}[t] \centering
    \includegraphics[width=0.85\textwidth]{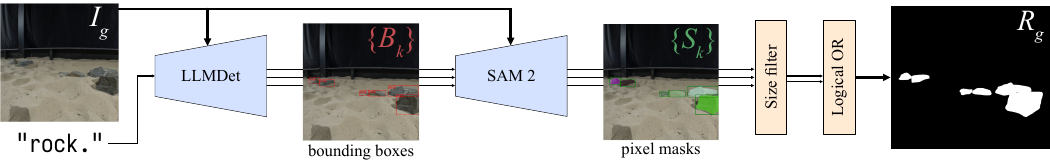}
    \caption{Full rock segmentation pipeline using LLMDet and SAM~2.}
    \label{fig:segmentation}
\end{figure*}

\section{Related work}
Here, we present an overview of current navigation and localisation techniques for planetary rovers, as well as works related to cross-view localisation and semantic segmentation.

\subsection{Planetary navigation and localisation}
Planetary localisation has traditionally relied on visual odometry from stereo reconstruction and matching \autocite{kou.etal.2025_crosssite,andolfo.etal.2023_precise}, and feature-based map matching techniques \autocite{ebadi.etal.2022_autonomous,EE.2018_rover} to estimate rover position. As stereo cameras and SLAM methods offer relative localisation, they are prone to drift over time \autocite{verma.etal.2024_enabling}. Recent efforts have also explored matching rover imagery to aerial or satellite maps for absolute localisation \autocite{verma.etal.2024_enabling,hook.etal.2022_topographical,dinsdale.etal.2022_absolute}. These, however, often require rich geometry via stereo or depth, or handcrafted features, limiting their applicability in monocular setups.

\subsection{Deep-learning-based cross-view localisation}
Cross-view localisation using deep learning has gained traction in both terrestrial \autocite{zhang.etal.2024_mtgeo,shugaev.etal.2024_arcgeo,zhu.etal.2022_transgeo} and planetary \autocite{chen.etal.2024_metric,zhao.etal.2024_lunar,franchi.ntagiou.2022_planetary} domains. These works approach the problem via dual-encoder architectures, which encode ground and aerial images into a shared feature space, enabling localisation via feature similarity.

Methods like TransGeo \autocite{zhu.etal.2022_transgeo} apply these approaches to Earth data, while LunarCV \autocite{chen.etal.2024_metric} and others \autocite{zhao.etal.2024_lunar,franchi.ntagiou.2022_planetary} apply this to simulated lunar data, with \autocite{chen.etal.2024_metric} requiring the rover's bearing to be known. In \autocite{franchi.ntagiou.2022_planetary}, integrating these models into sequential localisation frameworks with traditional odometry data using particle filters is explored. Training these models on planetary data, however, remains challenging due to the scarcity of real images with ground-truth localisation data. 

\subsection{Semantic segmentation and object detection}
Semantic segmentation is widely used in planetary robotics for terrain understanding, hazard detection, and localisation \autocite{swan.etal.2021_ai4mars}. Rocks in particular serve as reliable visual landmarks on the Martian surface. Prior work has explored multi-class \autocite{zhang.etal.2024_s5mars,barrett.etal.2022_noahh} and rock-specific \autocite{wei.etal.2025_rocknet,liu.etal.2023_rockformer} segmentation using CNNs and transformers.

To bridge the domain gap between synthetic and real data, recent advances in vision foundation models such as LLMDet \autocite{fu.etal.2025_llmdet} for open-vocabulary object detection and SAM~2 \autocite{ravi.etal.2024_sam} for segmentation offer robust, domain-invariant feature extraction. These models, pre-trained on broad, large-scale data, allow generalisation across domains without requiring task-specific training.

\section{Method}
Our goal is to estimate a rover's pose in a local aerial map using ground-view, \ang{90} horizontal field-of-view images. We decompose this into three parts, highlighted in \cref{fig:hero}: applying semantic segmentation to handle the domain gap (\cref{sec:semantic-seg}), training a dual-encoder network to perform cross-view matching (\cref{sec:transgeo}), and integrating this into a particle filter for state estimation over time (\cref{sec:particle-filter}).

\subsection{Domain-robust input via rock segmentation}\label{sec:semantic-seg}
To address the domain gap, we use foundation models LLMDet \autocite{fu.etal.2025_llmdet} for object detection and SAM~2 \autocite{ravi.etal.2024_sam} for segmentation in our ground view images, as shown in \cref{fig:segmentation}.

For a ground-view image $I_g \in [0,1]^{W\times H\times C}$ and full-stop-separated prompt string $P$, the LLMDet network produces a set of $K$ bounding boxes $B_k= (x_1, y_1, x_2, y_2)_k$:
\begin{equation}
    \mathcal B_g = \mathrm{LLMDet}(I_g, P) = \{ B_k \}_{k=1}^K \text{.}
\end{equation}
Each box $B_k$ surrounds one identified object that matches the prompt. We use the prompt string $P = \text{``\texttt{rock.}''}$ to identify the rocks visible in our images.

These bounding boxes are then passed on to the SAM~2 network, which attempts to produce a segmented boolean mask $S_k \in \{0,1\}^{W\times H}$ for each box in $\mathcal B_g$:
\begin{equation}
    \mathcal S_g = \mathrm{SAM}(\mathcal B_g) = \{S_k\}_{k=1}^K \text{.}
\end{equation}

The masks are filtered by size, and then the overall boolean rock mask $R_g \in \{0,1\}^{W\times H}$ is created from the pixel-wise $(u,v)$ logical-OR combination across these masks
\begin{equation}
    R_{g,(u,v)} = \textstyle\bigvee_{k=1}^{K} S_{k,(u,v)} \text{.}
\end{equation}

\subsection{Cross-view embedding via dual-encoder network}\label{sec:transgeo}
\begin{figure} \centering
    \includegraphics[width=\columnwidth]{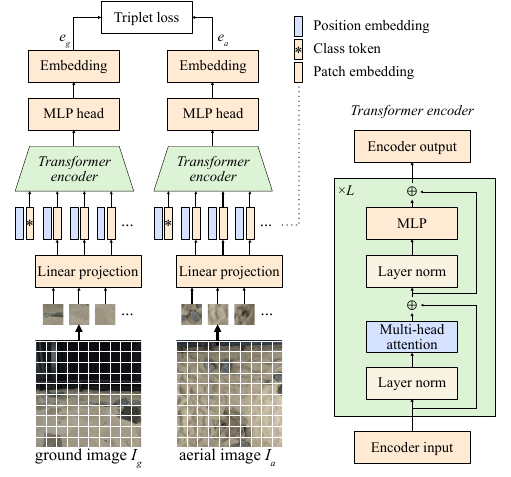}
    \caption{Dual-encoder cross-view localising network structure, adapted from \autocite{zhu.etal.2022_transgeo}. We use $L=12$ encoder blocks.}
    \label{fig:transgeo}
\end{figure}

We use a network architecture based on stage~1 of TransGeo~\autocite{zhu.etal.2022_transgeo} as our cross-view localising backbone, shown in \cref{fig:transgeo}. The network follows a dual-encoder structure, with separate learnable weights for the ground-view and aerial-view images. In their respective branches, the input images $I \in [0,1]^{W \times H \times C}$ for $H=W=256$, $C=3$ are divided into patches. These are passed alongside learnable positional encodings and a class token into a transformer encoder block. A fully-connected layer then produces the embeddings $e \in \mathbb R^{D}$ of dimension $D = 32$, which are normalised so that $\norm{e}_2 = 1$.

The network is trained using a soft-margin triplet loss \autocite{hu.etal.2018_cvmnet}. Our datasets consist of $N$ matched image pairs $\{(I_g, I_a)_n\}_{n=1}^N$. For a given query ground image $I_{g,i}$, its positive matching aerial image $I_a^+ = I_{a,i}$ and one negative non-matching aerial image $I_a^- = I_{a,n\neq i}$ are found. The triplet loss is then calculated as
\begin{equation}\label{eq:triplet-loss}
    \mathcal L_\text{triplet} = \log\left( 1 + e^{\alpha(d^+ - d^-)} \right)
\end{equation}
for
\begin{equation}
    d^+ = \norm{e_g, e_a^+}_2^2 \text{, } d^- = \norm{e_g, e_a^-}_2^2 \text{.}\label{eq:distances}
\end{equation}
This loss is exhaustively applied to each of the $2B(B-1)$ triplets -- including analogous triplets of the form $(I_a, I_g^+, I_g^-)$ -- in a training batch of $B$ pairs.

As we are considering the circumstance where $I_g$ has a \ang{90} horizontal field of view, we define $I_a$ and $I_g$ so that the camera location of $I_g$ is the lower-left corner of $I_a$, and the forwards direction of $I_g$ runs along the diagonal towards the centre of $I_a$.

The same network structure and training process is used for localising using the segmented rock masks, with segmentation outputs $R_g$ and $R_a$ used in place of $I_g$ and $I_a$.

\subsection{Particle filtering for state estimation}\label{sec:particle-filter}
The rover is localised over time with a particle filter \autocite{thrun.etal.2005_probabilistic}, relative to an overall geo-referenced local aerial map image $I_A$ (or corresponding rock mask $R_A$). The filter uses outputs from our networks as its measurement model.

The state space $\vec x_t = [x, y, \beta]$ is used for the state of our rover at time $t$, comprising the rover's lateral position $x$, $y$ and bearing angle $\beta$. Hence, particles $\vec x_t^{(m)}$ with weights $w_t^{(m)}$ for $m = 1,\ldots,M$ are tracked.

Prediction uses an approximately constant velocity model, allowing for variations in bearing. At each prediction step,  new bearings and distances
\begin{equation}
    \beta_t^{(m)} \sim \mathrm N(\beta_{t-1}^{(m)}, \sigma_\beta^2) \text{ and } d_t^{(m)} \sim \mathrm N(\mu_d, \sigma_d^2)
\end{equation}
are sampled, and then the state is propagated as
\begin{equation}
    \vec x_t^{(m)} = [x_{t-1}^{(m)} + d_t^{(m)} \sin \beta_t^{(m)}, y_{t-1}^{(m)} + d_t^{(m)} \cos \beta_t^{(m)}, \beta_t^{(m)}] \text{.}
\end{equation}

The observation at each time step is a new ground image $I_{g,t}$. To re-weight the particles, the localised aerial patch $I_{a,t}^{(m)}$ is extracted from $I_A$ for each particle. $I_{a,t}^{(m)}$ is aligned to the particle state $\vec x_t^{(m)}$ as with $I_g$ in \cref{sec:transgeo}. The embeddings $e_{g,t}$ and $e_{a,t}^{(m)}$ are calculated, and the cosine similarity $s_t^{(m)}$ between them is found. The weights are then calculated as 
\begin{equation}
    w_t^{(m)} = b^{s_t^{(m)}}
\end{equation}
for base $b$ before re-normalisation. Resampling is employed to avoid sample impoverishment.

\section{Dataset}
To investigate this cross-view localisation method, we have created a cross-view dataset from images and motion data captured with a rover in the Extraterrestrial Environment Simulation (Exterres) laboratory at the University of Adelaide. This is complemented by a higher-volume synthetic dataset designed to appear similar to the laboratory environment, generated using PANGU~\autocite{martin.dunstan.2021_pangu}. The combined datasets are available at \href{https://doi.org/10.5281/zenodo.17364038}{https://doi.org/10.5281/zenodo.17364038}, and examples are presented in the supplementary video.

\subsection{Laboratory-collected data}
The laboratory dataset was collected in a rectangular sandpit of approximately \qtyproduct{3 x 5}{\meter}. The pit has blackout curtains around all four sides and ceiling, and is lit by ambient lights in the top corners. A mix of real and foam prop rocks are distributed across the surface, to serve as the primary features for localisation.

\begin{figure} \centering
    \includegraphics[width=0.7\columnwidth]{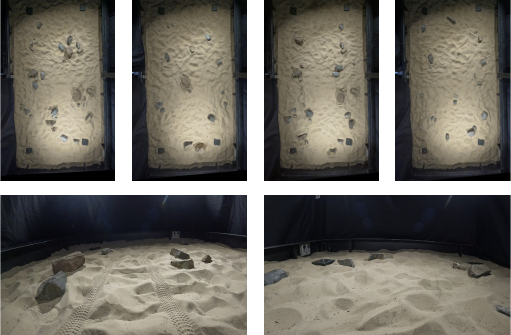}
    \caption{Examples of ground view (bottom) and rectified aerial view (top) images in our planetary analogue dataset.}
    \label{fig:dataset-examples}
\end{figure}

Six different rock configurations were set up, and then for each, an aerial still RGB image was collected. A Leo Rover \autocite{leorover.2025_leo} with an attached camera was then driven remotely across the surface in multiple trajectories, and ground-view RGB video was captured. While the rover was driving, an OptiTrack infra-red motion capture system \autocite{optitrack.2025_motion} was capturing the six-degree-of-freedom position and orientation of a cluster of IR markers attached to the rover. The tracked locations of additional IR markers in the corners of the pit, together with checkerboard calibration of the aerial camera, enabled the aerial images to be rectified and geo-referenced to the rover motion capture data. A few sample images are shown in \cref{fig:dataset-examples}, and further statistics about the dataset are presented in \cref{tab:dataset-details}.

\begin{table}
    \centering
    \caption{Dataset statistics summary.}
    \begin{tabular}{@{}lr@{}}
        \toprule
        Description & Value \\
        \midrule
        \multicolumn{2}{@{}c@{}}{\emph{Laboratory dataset}} \\
        Rock configurations & 6 \\
        Traverses & \\
        ---per configuration & 3 to 9 \\
        ---total & 40 \\
        Traverse duration & \\
        ---average & \SI{32}{\second} \\
        ---total & \SI{20}{\minute} \SI{58}{\second} \\
        Camera hardware & GoPro HERO10 Black \\
        Aerial image resolution & \numproduct{5568 x 4176} \\
        Ground video & \\
        ---resolution & \numproduct{3840 x 2160} \\
        ---horizontal field of view\footnotemark & \ang{120} \\ 
        ---capture rate & \SI{30}{\hertz} \\
        Motion capture refresh rate & \SI{120}{\hertz} \\
        \midrule
        \multicolumn{2}{@{}c@{}}{\emph{Synthetic dataset}} \\
        Scenes & 10 \\
        Image pairs per scene & 500 \\
        Aerial image resolution & \numproduct{512 x 512} \\
        Ground image & \\
        ---resolution & \numproduct{512 x 512} \\
        ---horizontal field of view & \ang{90} \\
        \bottomrule
    \end{tabular}
    \label{tab:dataset-details}
\end{table}

\subsection{Synthetic data}
The synthetic dataset, created using the PANGU software \cite{martin.dunstan.2021_pangu}, comprises pairs of aerial and ground images of procedurally generated terrain and rocks, tuned to produce output that resembles the laboratory-collected data. The synthetic scenes, like the real-world sandpit, are limited to a rectangular area. Fractal algorithms built into PANGU generate a height map and a colour albedo map for the base terrain of each scene. Additional built-in algorithms generate and place rocks within the scenes, following prescribed size, colour, depth, and shape distributions.

Ten scenes are generated, each initialised with different random seeds. In each scene, a set of random rover poses are sampled across the terrain, and RGB images of a ground view and corresponding aerial view are captured. The lighting configuration is varied for each image pair, with sun azimuth and elevation angles sampled uniformly from $[\ang{0}, \ang{360})$ and $[\ang{40}, \ang{60}]$ respectively. Further details can again be found in \cref{tab:dataset-details}.

\footnotetext{The original video is \ang{120} horizontal field of view, but we crop to the central \ang{90} for our experiments.}

\section{Experiments}
In this section, we outline the experiments and results for the different elements of our cross-view localisation method.

\subsection{Domain-robust rock segmentation}
\begin{figure} \centering
    \includegraphics[width=0.7\columnwidth]{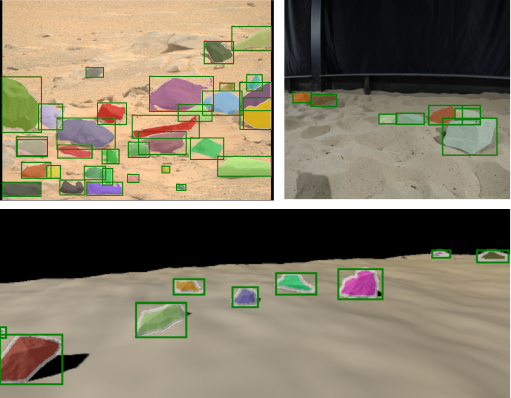}
    \caption{Qualitative examples showing the success of the LLMDet + SAM~2 segmentation on a real Mars image (top left), real planetary analogue image (top right), and synthetic image (bottom).}
    \label{fig:qualitative-segmentation}
\end{figure}

First, we demonstrate the ability of foundation models to segment rocks from our images. \Cref{fig:qualitative-segmentation} shows qualitative examples of LLMDet + SAM~2 successfully addressing the domain gap by segmenting rocks across data domains.

A ground-truth set of segmented rock masks was constructed by manually guiding SAM~2. Images from three of the ground-truth real data trajectories ($N=96$) were segmented from manually selected points describing each visible rock.

To evaluate the combination of LLMDet's object detection with SAM~2 against manual prompting, the full segmentation pipeline in \cref{fig:segmentation} was run on the same images to create a prediction set of masks. LLMDet was configured with box and text thresholds $t_b = t_t = 0.2$. This pipeline is also compared against using SAM~2 alone, where a uniform grid of points is used to guide the segmenter to attempt to find key objects in the images. SAM~2 used a \numproduct{32 x 32} point grid, with IoU threshold 0.95, stability score threshold 0.97, box non-maximum suppression threshold 0.95, and zero crop layers.

The pixel-level prediction statistics are summarised in \cref{tab:segmentation-results}. These demonstrate that the use of LLMDet with SAM~2 over SAM~2 alone is highly advantageous, and that this model has high precision at the cost of weaker recall. This is useful, as preliminary experiments indicated that our neural network was much more sensitive to false positives than false negatives in real-world localisation.

\begin{table}\centering
    \caption{Automatic rock mask prediction statistics, compared against ground truth manually guided SAM~2 masks.}
    \label{tab:segmentation-results}
\begin{tabular}{@{}lrrrr@{}}
\toprule
               & \multicolumn{2}{c}{Precision} & \multicolumn{2}{c}{Recall} \\ \cmidrule(lr){2-3} \cmidrule(l){4-5} 
Automatic method         & Mean        & Std.       & Mean       & Std.     \\ \midrule
LLMDet + SAM 2 & 0.934       & 0.063           & 0.671      & 0.287         \\
SAM 2          & 0.770       & 0.189           & 0.443      & 0.257         \\ \bottomrule
\end{tabular}
\end{table}

\begin{table*}\centering
    \caption{Validation results of the cross-view localising neural network backbone. Details top-$k$ matching results, as well as the mean similarity separation $d^+ - d^-$ as in \cref{eq:distances}, where higher is better.}
    \label{tab:nn-validation}
\begin{tabular}{@{}llrrrrr@{}}
\toprule
                             &                              & \multicolumn{4}{@{}c@{}}{Validation image-matching results}                                     & \multicolumn{1}{l}{}       \\ \cmidrule(lr){3-6}
Training data                & Validation data              & top-1                & top-5                & top-1\%              & top-20\%             & Mean similarity separation \\ \midrule
Synthetic RGB ($N = 4500$)   & Synthetic RGB ($N = 1000$)   & 88.0\%               & 98.0\%               & 99.0\%               & 100.0\%              & 0.8698                     \\
                             & Real RGB ($N = 676$)         & 1.2\%                & 4.9\%                & 6.2\%                & 46.9\%               & 0.1263                     \\
Real RGB ($N = 682$)         & Real RGB ($N = 676$)         & 5.2\%                & 19.8\%               & 22.2\%               & 90.2\%               & 0.6007                     \\
Synthetic RGB + Real RGB     & Real RGB ($N = 676$)         & 16.1\%               & 56.8\%               & 62.7\%               & 99.4\%               & 0.6578                     \\
Synthetic masks ($N = 4500$) & Synthetic masks ($N = 1000$) & 66.9\%               & 87.1\%               & 91.5\%               & 99.7\%               & 0.7938                     \\
                             & Real masks ($N = 96$)       & 14.6\% & 49.0\% & -- & 80.2\% & 0.3430   \\ \bottomrule
\end{tabular}
\end{table*}

\subsection{Learning dual-encoder cross-view embeddings}
Next, we train and evaluate our network backbone on different combinations of training and validation data. Throughout this section, we consider a synthetic and a real dataset. The synthetic dataset comprises PANGU image pairs ($N_\text{train} = 4000$, $N_\text{val}=1000$) at random poses. The real dataset comprises image pairs ($N_\text{train} = 682$, $N_\text{val} = 676$) extracted from real rover trajectories. The six real data rock configurations were divided into training and validation sets, and the corresponding image pairs from two runs in each configuration were extracted, sampled from the ground video at \SI{1}{\hertz}. The validation results of matching aerial images to ground images are presented in table \cref{tab:nn-validation}.

Each of these networks was trained to 500 epochs with a batch size of 32. We use the ASAM optimiser \autocite{kwon.etal.2021_asam} with a learning rate of \num{1e-5} and cosine scheduling. The weight of the soft-margin triplet loss as in \cref{eq:triplet-loss} is set to $\alpha = 20$. The model weights are initialised as in \autocite{zhu.etal.2022_transgeo}.

The first network is trained on our synthetic RGB image pairs. The results indicate that, like previous works, the network is able to learn to localise in synthetic space-domain data (top-20\% matching = 100\%). They also show that the network trained on synthetic data does not successfully generalise to also localise in real data (top-20\% = 46.9\%), validating our concern with the domain gap problem.

Next a network is trained on real data. Note that the image pairs used in the real data training set are sampled from trajectories and hence have a much smaller viewpoint difference between image pairs, so the overall matching scores are lower. We find that a network trained on real data only has reasonable matching performance (top-20\% = 90.2\%), but a second network trained on synthetic data and then fine-tuned for a further 500 epochs on real data has considerably better performance (top-20\% = 99.4\%).

Finally, we consider a network trained only on masked synthetic images. To help this network be robust to noise, the training data for the synthetic masks was produced with SAM~2's grid initialisation. This network successfully matches images in the synthetic masked validation set (top-20\% = 99.7\%). Evaluating this network on the ground-truth masked images demonstrates its successful domain adaptation (top-20\% = 80.2\%).

\begin{figure} \centering
    \includegraphics[width=\columnwidth]{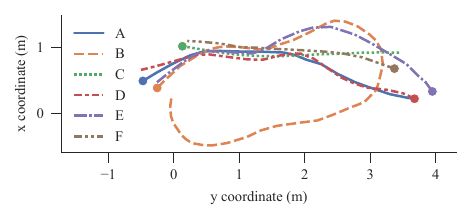}
    \caption{The six trajectories, A to F, used in the validation set for evaluating our particle filter. The start point of each is marked with a circle.}
    \label{fig:trajectories}
\end{figure}

\subsection{Particle filtering for state estimation}
Finally, we evaluate our neural network backbone as the measurement model for state estimation in a particle filter.

\subsubsection{Real data inference}
To evaluate the localisation performance, we consider the six trajectories (runs A to F) that formed the validation partition of the real dataset, plotted in \cref{fig:trajectories}. For each trajectory we run a particle filter with $M=500$ particles. We sample images from the ground-view video at \SI{1}{\hertz}, and discard images that differ from the previous by a distance of less than \SI{0.1}{\meter}. We condition our process model with $\sigma_\beta = \ang{15}$, $\mu_d = \SI{0.25}{\meter}$, and $\sigma_d = \SI{0.1}{\meter}$. Our particles are similarly initialised in a cloud around the starting state with $\sigma_\beta = \ang{15}$ and $\sigma_d = \SI{0.1}{\meter}$. Our measurement model uses base $b=\num{1e10}$.

\begin{figure}\centering
    \includegraphics[width=\columnwidth]{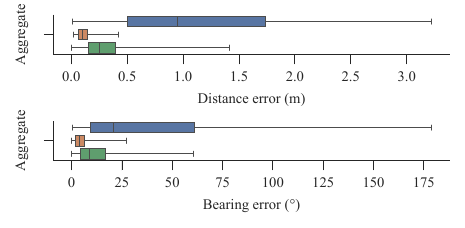}
    \includegraphics[width=\columnwidth]{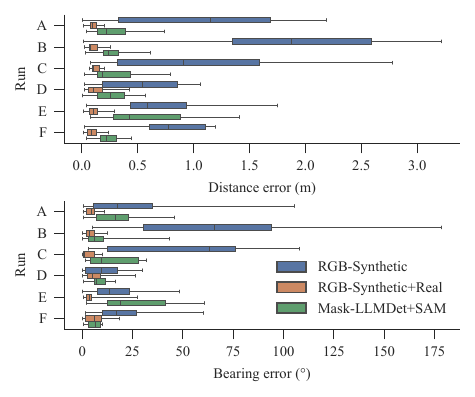}
    \caption{Comparison of the distribution of particle filter range and bearing errors at each timestep in aggregate and separated by run.}
    \label{fig:pf-errors}
\end{figure}

We start by considering three experiments -- RGB data with the network trained only on synthetic data (``RGB-Synthetic''), RGB data with the network trained on synthetic and fine-tuned on real data (``RGB-Synthetic+Real''), and masked data with the network trained on synthetic data only and the real data semgented using LLMDet and SAM~2 (``Mask-LLMDet+SAM''). Note that the data used to fine-tune the second network does not contain any images from the rock configurations used for runs A to F. The distribution of range and bearing errors compared to the ground truth at each timestep in aggregate and by run is shown in \cref{fig:pf-errors}.

These results indicate that the localising network performs best when it is trained on real-world representative data, bypassing the domain gap. That said, our method of using semantic segmentation to overcome the domain gap allows for successful localisation in instances where real-world data is not available. Our method has a considerably lower median and maximum distance error than the synthetic-only no-domain-adaptation method across all six runs, and a similar trend for bearing error across all but one (run E).

\subsubsection{Automatic versus manual segmentation}
\begin{figure}\centering
    \includegraphics[width=\columnwidth]{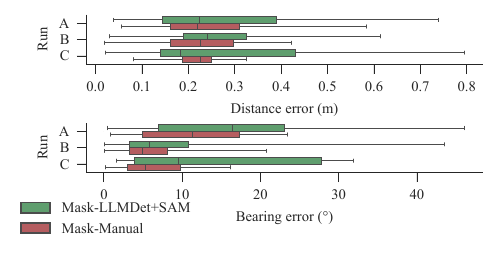}
    \caption{Comparison of particle filter errors between automatic and manual semantic segmentation of ground-view images.}
    \label{fig:pf-sam-manual}
\end{figure}

We also compare the inference data from the LLMDet and SAM~2 segmentation pipeline against our manual ground-truth segmentation. The mask-trained network was evaluated on both image sets for three runs, with results shown in \cref{fig:pf-sam-manual}. Clearly the manual segmentation consistently performs better, but the difference in median distance error is small.

\begin{figure}\centering
    \includegraphics[width=\columnwidth]{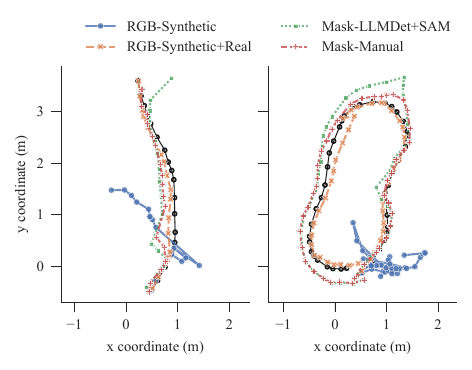}
    \caption{Qualitative examples of particle filter runs A (left) and B (right). Black line and circles indicate ground truth location.}
    \label{fig:pf-qualitative}
\end{figure}

Two qualitative examples of the localisation runs across the different methods are shown in \cref{fig:pf-qualitative}. These results particularly highlight the importance of domain gap adaptation, as the RGB-Synthetic network runs diverge completely. They also demonstrate that all three other approaches are able to track the rover's state successfully, but that slightly larger local errors occur at times with the mask networks compared to the fine-tuned RGB network.

\begin{figure} \centering
    \includegraphics[width=\columnwidth]{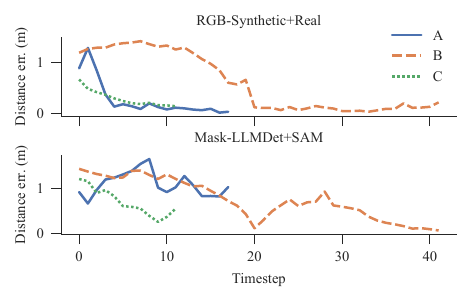}
    \caption{Distance error over time for known-bearing, unknown-location particle initiation across runs A to C.}\label{fig:lostinspace}
\end{figure}

\subsubsection{Lost-in-space localisation}
We also investigate the case of lost-in-space particle initiation, where the rover's starting bearing is known, but its position is unknown. In these experiments, the particles are initialised in an evenly-spaced grid across the sandpit surface. These results are shown in \cref{fig:lostinspace}. As can be seen, for all three evaluated runs, the real-image fine-tuned network is able to successfully localise the rover after some initialisation period. The LLMDet+SAM network shows reasonable performance for two out of three runs, but diverges for the third (run A).

\subsection{Compute resources}
All of these experiments were run using a single RTX 2080~Ti GPU. This is more powerful than typical space-qualified compute hardware at present, but as the demand for autonomy and interest in machine learning continues to increase, more advanced compute modules are becoming space qualified \autocite{Felix2024total}.

On this hardware, the segmentation of a ground-view image using LLMDet and SAM~2 takes a mean of \SI{811}{\milli\second}, with a standard deviation of \SI{190}{\milli\second}. The inference with our TransGeo-based network backbone takes a mean of \SI{30}{\milli\second} per particle, with a standard deviation of \SI{2.8}{\milli\second}. This results in an approximate \SI{15}{\second} processing time for each ground-view image. This does exceed the \SI{1}{\hertz} at which the ground-view images were sampled, but this is a small-scale experiment and real rovers typically drive much more slowly. The frequency of the ground image sampling could also be reduced, especially if this absolute localisation method were to be used as intermittent correction for traditional relative odometry methods.

\section{Conclusions}
We successfully demonstrate the applicability of a dual-encoder cross-view localisation network to a planetary rover localising itself in a local aerial image. Specifically, we have investigated the challenging setting of monocular RGB images with a limited \ang{90} horizontal-field-of-view. We use synthetic data and semantic segmentation with vision foundation models to explicitly address the domain gap between synthetic and real data, and demonstrate a successful method of localisation on real data that does not require any real labelled data at train time. We also contribute a dataset of labelled synthetic image pairs and ground-truth-localised rover trajectories with corresponding georeferenced aerial images.

Valuable future work would include reducing network size and complexity for better inference times on space-grade hardware, and integrating this absolute localisation method with traditional odometry-based relative localisation.

\AtNextBibliography{\small}
\printbibliography

\end{document}